\documentclass{article}
\usepackage{ijcai22}

\usepackage{times}

\usepackage{soul}
\usepackage{url}
\usepackage[hidelinks]{hyperref}
\usepackage[utf8]{inputenc}
\usepackage[small]{caption}
\usepackage{graphicx}
\usepackage{amsmath}
\usepackage{amssymb} 
\usepackage{multirow}
\usepackage{booktabs}
\usepackage{algorithm}
\usepackage{algorithmic}
\usepackage{subfigure}
\usepackage{arydshln}
\title{AdMix: A Mixed Sample Data Augmentation Method for Neural Machine Translation}

\author{
Chang Jin\footnote{Contact Author}\and
Shigui Qiu\and
Nini Xiao\and
Hao Jia\\
\affiliations
Institute of Aritificial Intelligence, School of Computer Science and Technology, Soochow University\\
\emails
\{cjin, sgqiu, nnxiaoxiao, hjia\}@stu.suda.edu.cn,
}

\begin{document}

\maketitle

\begin{abstract}
In Neural Machine Translation (NMT), data augmentation methods such as back-translation have proven their effectiveness in improving translation performance.
In this paper, we propose a novel data augmentation approach for NMT, which is independent of any additional training data. Our approach, {\it AdMix}, consists of two parts: 1) introduce faint discrete noise (word replacement, word dropping, word swapping) into the original sentence pairs to form augmented samples; 2) generate new synthetic training data by softly mixing the augmented samples with their original samples in training corpus.
Experiments on three translation datasets of different scales show that AdMix achieves significant improvements (1.0 to 2.7 BLEU points) over strong Transformer baseline. When combined with other data augmentation techniques (e.g., back-translation), our approach can obtain further improvements.
\end{abstract}

\begin{figure*}[htbp]
    \flushleft
    \centerline{\includegraphics[width=16cm]{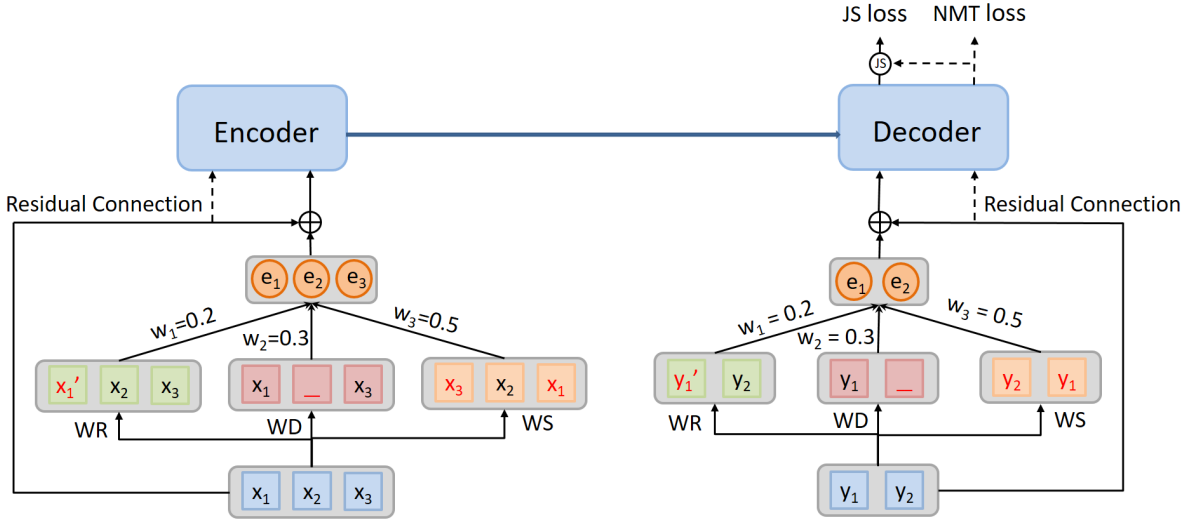}}
    \caption{Illustration of the proposed method, {\it AdMix}, which linearly interpolates the original sentence with its discrete transformed sentence to generate synthetic training sample. The dashed line represents the data flow of the original sentence, the solid line represents the data flow of the synthetic sentence.}\label{fig:fig1}
\end{figure*}

\section{Introduction}

Data augmentation are techniques used to increase the amount of data by adding slightly modified copies of already existing data or newly created synthetic data from existing data~\cite{li2021data}. These methods can significantly boost the accuracy of deep learning methods. For image classiﬁcation tasks, there are various data augmentation methods like random rotation, mirroring, cropping and cut-out~\cite{krizhevsky2012imagenet,devries2017improved}.
However, universal data augmentation techniques in natural language processing (NLP) tasks have not been thoroughly explored due to the discrete nature of natural language~\cite{wei2019eda,gao2019soft}. 

According to the diversity of augmented data, Li et al.~\shortcite{li2021data} frame data augmentation methods into three categories, including paraphrasing, noising, and sampling. Specifically, the purpose of the paraphrase-based methods is to generate augmented samples that are semantically similar to the original samples. Back-translation~\cite{sennrich-etal-2016-improving} is a typical paraphrase-based method to translate monolingual data by an inverse translation model. While effective, back-translation method becomes inapplicable when target-side monolingual data is limited. The sampling-based methods master the distribution of the original data to sample novel data points as augmented data. For this type of methods, a common approach is to use the pretrained language model to generate labeled augmented sentences~\cite{anaby2020not,kumar-etal-2020-data}. The last category is based on noising, which adds discrete or continuous noise under the premise of guaranteeing validity. The noising-based methods consist of word dropping, word replacement, word swapping and Mixup, etc. Due to their simplicity and effectiveness, we focus on these methods in this paper. Nevertheless, the previously proposed noise-based methods have their limitations. For example, injecting faint discrete noise (including word replacement, word dropping and word swapping) into sentences produces sentences with limited diversity. Guo et al.~\shortcite{guo-etal-2020-sequence} propose a sequence-level variant of Mixup~\cite{Zhang2017mixup}. Despite its effectiveness, this method is less interpretable and more difficult than the discrete noising-based methods.

In this paper, we propose a new data augmentation method for neural machine translation task. Our approach, {\it AdMix}, first introduces an appropriate amount of discrete noise to obtain augmented samples, and then mixes these augmented samples by randomly sampling from the Dirichlet($\alpha$, . . . , $\alpha$) distribution. Once these augmented sentences are mixed, we use a residual connection to combine the mixed samples with their original samples through a second random convex combination sampled from a Beta($\beta$, $\beta$) distribution. In this way we obtain the final synthetic samples. 
To verify the effectiveness of our method, we conduct experiments on three machine translations tasks, including IWSLT14 German to English, LDC Chinese to English, and WMT14 English to German translation tasks. The experimental results show that our method yields significant improvements over strong Transformer baselines on datasets of different scales. 

We highlight our contributions in three aspects:
\begin{itemize}
\item We propose a simple but effective strategy to boost NMT performance, which linearly interpolates the original sentence with its discrete transformed sentence to generate new synthetic training data.
\item Experiments on three translation benchmarks show that our approach achieves significant improvements over various strong baselines.
\item Our method can combine with other data augmentation techniques to yield further improvements.
\end{itemize}

\section{Related Work}
A large number of data augmentation methods have been proposed recently, and we will present several related works on data augmentation for NMT. 

One popular study is back-translation~\cite{sennrich-etal-2016-improving} which generates synthetic samples based on the conditional probability distribution of the reverse translation model. But back-translation requires training a reverse translation model, which usually consumes giant computing resources. Another similar method is self-training~\cite{He2020Revisiting}, which augments the original labeled dataset with unlabeled data paired with the model’s prediction. Different from back-translation, it uses source-side monolingual data to generate pseudo-parallel sentences. Self-training avoids the need to train a reverse translation model, but it is also not suitable in the case of limited monolingual data.

The noising-based methods are also common data augmentation technologies. Such methods are not only easy to use, but also to improve the robustness of the model~\cite{Miyato2017Adversarial}. Artetxe et al.~\shortcite{artetxe2017unsupervised} randomly choose several words in the sentence and swap their positions. Lample et al.~\shortcite{lample2018unsupervised} randomly remove each word in the source sentence with probability $p$ to help train the unsupervised NMT model. Xie et al.~\shortcite{xie2017data} randomly replace the word with a placeholders token. Similarly to Xie et al.~\shortcite{xie2017data}, Wang et al.~\shortcite{wang-etal-2018-switchout} propose a method that randomly replaces words in both the source and target sentences with other words in the vocabulary. Although faint discrete noises are not strong enough to alter the meaning of the input sentence, they may result in the loss of lexical meaning of specific words. Thus, it affects the correspondence between source-side sentences and target-side sentences. Another problem with the discrete noising-based approach is the limited diversity of its operations. Cheng et al.~\shortcite{cheng-etal-2018-towards} simply add the Gaussian noise to all of word embeddings in a sentence to simulate possible types of perturbations. The formula is 
\begin{equation}
\begin{aligned}
\mathbf{E}\left[\mathbf{x}_{i}^{\prime}\right]=\mathbf{E}\left[\mathbf{x}_{i}\right]+\epsilon, \quad \epsilon \sim \mathbf{N}\left(0, \sigma^{2} \mathbf{I}\right),
\end{aligned}
\end{equation}
where the vector $\epsilon$ is sampled from a Gaussian distribution with variance $\sigma^{2}$. $\sigma$ is a hyperparameter. This method is also not guaranteed to maintain semantics.

In addition to the above methods, there is another data augmentation method that has attracted a lot of attention recently, called Mixup. The Mixup data augmentation technique was first presented in image classification by Zhang et al.~\shortcite{Zhang2017mixup}. Hendrycks et al.~\shortcite{hendrycks2020augmix} proposed an advanced Mixup method called AugMix, a data processing technique that mixed augmented images and enforced consistent embeddings of the augmented images. Inspired by the work, Guo et al.~\shortcite{guo-etal-2020-sequence} proposed a sequence-level variant of Mixup. As described by Guo et al.~\shortcite{guo-etal-2020-sequence}, a pair of training examples ($x_1$, $y_1$) and ($x_2$, $y_2$) are sampled from the training set. And the synthetic sentences are:
$$\hat{x} = \lambda x_{1} + (1-\lambda) x_{2},$$
$$\hat{y} = \lambda y_{1} + (1-\lambda) y_{2}.$$
$\lambda$ is drawn from a Beta($\beta$, $\beta$) distribution, which is controlled by the hyperparameter $\beta$. Different from Guo et al.~\shortcite{guo-etal-2020-sequence}, Cheng et al.~\shortcite{cheng-etal-2020-advaug} firstly build their adversarial samples, and then generate a new synthetic sample by interpolating between the adversarial samples. Although Cheng et al.~\shortcite{cheng-etal-2020-advaug} guarantee the invariance of the original sentences semantic information compared to Guo et al.~\shortcite{guo-etal-2020-sequence}, they need an additional language model (LM) to generate the adversarial samples. Compared with these studies, our method does not require additional resources, and the generated synthetic samples are more interpretable than Mixup method.

\section{AdMix}
In our approach, {\it AdMix}, the goal is to generate new synthetic training examples as illustrated in Figure \ref{fig:fig1}. It is applied to both the source and target sequences to build new synthetic samples, which are used as augmented translation pairs for training purposes. This means that our approach only influence the training process of NMT without changing its inference process. AdMix is divided into two stages: discrete noise and data mixing, and the details are described as follows.

\subsection{Discrete Noise}
In this stage, we introduce an appropriate amount of discrete noises to obtain several augmented samples. Let $X \in \mathbb{R}^{s \times V}$, $Y \in \mathbb{R}^{t \times V}$ represent a source sequence of length $s$ and a target sequence of length $t$, respectively. $V$ denotes the size of vocabulary. Given a pair of training example $(X, Y)$ in training set, we perform word replacement (WR), word swapping (WS), word dropping (WD) operations on the source and target sentences to obtain $(X_{wr}, Y_{wr})$, $(X_{ws}, Y_{ws})$, $(X_{wd}, Y_{wd})$ respectively. For example, we swap the positions of $x_{1}$ and $x_{3}$, replace $x_{1}$ with $x_{1}^{\prime}$, and remove $x_{2}$ in the sentence for WS, WR, WD respectively (see Figure \ref{fig:fig1}).

In practice, to ensure that the amount of noise is proportional to the sentence length $l$, we set a hyperparameter $\gamma$. For WS and WR operations, the number of changed words is $n=\gamma l$. For WD operation, we randomly remove each word in the sentence with probability $\gamma$.

\subsection{Data Mixing}
After obtaining the augmented sentences, we first obtain their embedding sequences separately: $(\mathbf{E}\left[\mathbf{X}_{wr}\right], \mathbf{E}\left[\mathbf{Y}_{wr}\right])$, $(\mathbf{E}\left[\mathbf{X}_{ws}\right], \mathbf{E}\left[\mathbf{Y}_{ws}\right])$, $(\mathbf{E}\left[\mathbf{X}_{wd}\right], \mathbf{E}\left[\mathbf{Y}_{wd}\right])$. Inspired by Hendrycks et al.~\shortcite{hendrycks2020augmix}, we choose to use elementwise convex combinations to mix them and the coefﬁcient vector is randomly sampled from a Dirichlet($\alpha$, . . . , $\alpha$) distribution. The word embedding of these augmented data after softly mixing is $(\mathbf{E}\left[\mathbf{X}_{ad}\right], \mathbf{E}\left[\mathbf{Y}_{ad}\right])$. Finally, we use a residual connection to combine the mixed samples with their original samples by sampling $m$ from a Beta($\beta$, $\beta$) distribution, and the word embedding of the final synthetic sentence is
\begin{equation}
\mathbf{E}\left[\mathbf{X}_{admix}\right]=m \mathbf{E}\left[\mathbf{X}_{ad}\right]+ (1-m) \mathbf{E}\left[\mathbf{X}\right],
\end{equation}
\begin{equation}
\mathbf{E}\left[\mathbf{Y}_{admix}\right]=m \mathbf{E}\left[\mathbf{Y}_{ad}\right]+ (1-m) \mathbf{E}\left[\mathbf{Y}\right],
\end{equation}
where $\mathbf{E}\left[\mathbf{X}\right],\mathbf{E}\left[\mathbf{Y}\right]$ denote the word embedding at the source and target sides of original sentence pairs respectively.

\begin{algorithm}[tb]
\caption{AdMix Pseudocode}
\label{alg:algorithm}
\begin{algorithmic}[1] 
\STATE \textbf{Input}:  Model $p$ , Loss $\mathcal{L}$, sentence $x_{orig}$, $y_{orig}$,  \text{\quad Operations =\{replace, drop , swap\}} \\
\STATE \textbf{function} AdMix($x_{orig},y_{orig},k = 3,\alpha=1,\beta=1$):\\
\STATE \quad Fill $x_{ad}$, $y_{ad}$ with zeros
\STATE \text{\quad Sample mixing weights ($w_1$,$w_2$,...,$w_k$) $\sim$ }
\text{\qquad Dirichlet($\alpha$, $\alpha$, . . . , $\alpha$)}
\STATE \quad $\textbf{for}$ {each $i \in [1,$k$]$} 
\STATE \qquad $x_{noise}^i$ = Operations[i]($x_{orig}$)
\STATE \qquad $y_{noise}^i$ = Operations[i]($y_{orig}$)
\STATE \qquad $x_{ad} += w_i \cdot$ Embedding($x_{noise}^i$)  
\STATE \qquad $y_{ad} += w_i \cdot$ Embedding($y_{noise}^i$) 
\STATE \quad $\textbf{end for}$
\STATE \quad $x_{orig}$ = Embedding($x_{orig}$)
\STATE \quad $y_{orig}$ = Embedding($y_{orig}$)
\STATE \quad Sample weights m $\sim$ Beta($\beta$,$\beta$)
\STATE \quad Interpolate $x_{admix} = mx_{ad} + (1-m)x_{orig}$
\STATE \quad Interpolate $y_{admix} = my_{ad} + (1-m)y_{orig}$
\STATE \quad \textbf{return} $x_{admix}$, $y_{admix}$
\STATE \textbf{end function}
\STATE $x_{admix}$, $y_{admix}$ = AdMix($x_{orig}$,$y_{orig}$)
\STATE \textbf{Loss output:} \text{$\mathcal{L}(p(y|x_{orig};y_{orig}))$ + $\lambda$ JS$(p(y|x_{orig};y_{orig})$;}
\text{\quad $p(y|x_{admix};y_{admix}))$}
\end{algorithmic}
\end{algorithm}

\begin{table*}[ht]
\renewcommand\tablename{Table}
\centering
\begin{tabular}{l||cccccc||c||c}
\hline
  \multirow{2}{*}{Method} &\multicolumn{6}{c||}{Zh-En} &\multirow{2}{*}{En-De} & \multirow{2}{*}{De-En}\\
    \cline{2-7} & \small{NIST02} & \small{NIST03} & \small{NIST04} & \small{NIST05} & \small{NIST08} & \small{AVG} & \\
\hline\hline
Transformer & 47.48 & 46.77 & 47.92 & 47.02 & 37.99 & 45.44  & 27.30 &34.43\\
Swap     & 48.46 & 47.01 & 48.51 & 47.86 & 37.71 & 45.91  & 27.48 &34.65\\
WordDrop & 48.00 & 47.05 & 48.28 & 47.35 & 38.06 & 45.75  & 27.55 &35.03\\
Switchout& 48.02 & 47.96 & 48.06 & 48.73 & 39.07 & 46.37  & 27.60 &35.04\\
SeqMix   & 48.37 & 47.23 & 48.41 & \textbf{48.79} & 37.63 & 46.11  & 28.10 &35.49\\
AdMix   & \textbf{48.98} & \textbf{48.41} & \textbf{49.62} & 48.38 & \textbf{40.32} & \textbf{47.14}  & \textbf{28.26} & \textbf{37.10}\\
\hline
\end{tabular}
\caption{BLEU scores on LDC Zh-En, WMT En-De, and IWSLT De-En translation.}\label{tab:tab1}
\end{table*}

\subsection{Training Objectives}
The pseudocode of AdMix is shown in Algorithm \ref{alg:algorithm}. We couple this augmentation scheme with a loss that enforces a consistent embedding by the model across diverse augmentations of the same input sentences. Since the semantic meanings of the sentences are approximately preserved after AdMix operation, we can incorporate Jensen-Shannon divergence consistency loss into the training objective by encouraging the model to make similar predictions between original samples and synthetic samples. For this purpose, we minimize the Jensen-Shannon divergence among the posterior distributions of the original sample $(x, y)$ and its augmented variants $(x_{admix}, y_{admix})$. The training objective can be written as:
\begin{equation}
\mathcal{L}  = \mathcal{L}_{ce}(x,y) + \lambda JS(p_{orig};p_{admix}),
\end{equation}
\begin{equation}
\mathcal{L}_{\text {ce}}=-\sum_{i=1}^{T} \log P\left(y_i \mid x ; y_{<i} \right),
\end{equation}
where $\mathcal{L}_{ce}$ denotes the standard cross-entropy loss applied to the original samples, and $T$ refers to the length of the target sentence. $JS$ denotes Jensen-Shannon divergence. $p_{orig}$ and $p_{admix}$ are the probability distributions of model by respectively feeding the original sentence and the augmented sentence. $\lambda$ is a hyperparameter, which balances the importance between two loss function. 

The Jensen-Shannon divergence can be understood to measure the average information that the sample reveals about the identity of the distribution from which it was sampled~\cite{hendrycks2020augmix}. This loss can be computed by 
\begin{equation}
JS(p_{orig};p_{admix}) = 1/2 (KL[p_{orig}||M] + KL[p_{admix}||M]),
\end{equation}
where $M = (p_{orig} + p_{admix}) / 2$.
Although KL-divergence has been widely adopted as the divergence metric in previous works~\cite{Miyato2017Adversarial,clark-etal-2018-semi,xie2020unsupervised}, it has been shown that the JS divergence loss can endow the model with more stability and consistency across a diverse set of inputs~\cite{kannan2018adversarial,hendrycks2020augmix}. 
Therefore, we utilize the Jensen-Shannon (JS) divergence consistency loss as our loss function in this paper.

\section{Experiments}
We conduct experiments on the following machine translation tasks to evaluate our method: LDC Chinese-English (Zh-En), IWSLT14 German-English (De-En), and WMT14 English-German (En-De). The performance is evaluated with the 4-gram BLEU score~\cite{papineni2002bleu} calculated by the multi-bleu.perl script. For De-En and En-De, we report case-sensitive tokenized BLEU scores, while for Zh-En, we report case-insensitive tokenized BLEU scores. 

\subsection{Setup}
\textbf{Datasets:} For IWSLT14 German-English, following Edunov et al.~\shortcite{edunov-etal-2018-classical}, we apply the byte-pair encoding (BPE)~\cite{sennrich-etal-2016-neural} script to preprocess the training corpus with 10K joint operations, which consists of 0.16M sentence pairs. The validation set is split from the training set and the test set is the concatenation of tst2010, tst2011, tst2012, dev2010, and dev2012. For the Chinese-English translation task, the training set is the LDC corpus which contains 1.25M sentence pairs. The validation set is the NIST 06 dataset, and test sets are NIST 02, 03, 04, 05, 08. We apply BPE for Chinese and English respectively, and the merge operations are both 32k. For English-German translation, we use the WMT14 corpus consisting of 4.5M sentence pairs. The validation set is newstest2013 and the test set is newstest2014. We build a shared vocabulary of 32K sub-words using the BPE script.

\noindent\textbf{Training Details:} We choose Transformer as our translation model. For IWSLT14 German-English tasks, the dimensions of the embedding, feed-forward network, and the number of layers of the Transformer models are 512, 1024, and 6 respectively. The dropout rate is 0.3, and the batch size is 8192 tokens. For LDC Chinese-English task and WMT14 German-English task, the dimensions of the embedding, feed-forward network, and the number of layers of the Transformer models are 512, 2048, and 6 respectively. The dropout rate is 0.3, 0.1 separately for Zh-En and En-De tasks, and the batch size is both 8192 tokens. We train on two V100 GPUs and accumulate the gradients 2 times before updating. For all models but except the En-De task model, we use Adam with learning rate 5 × $10^{-4}$ and the inverse sqrt learning rate scheduler to optimize the models. For En-De task model, we use Adam with learning rate 7 × $10^{-4}$ and the inverse sqrt learning rate scheduler. There are two important hyperparameters in our approach, $\lambda$ in objective loss function and the discrete noise fractions $\gamma$. For all datasets, we set the noise fractions $\gamma$ as 0.1 and the hyperparameter $\lambda$ as 10 by default.
we also explore the effect of noise fractions $\gamma$ and hyperparameter $\lambda$ on the validation set, and then we choose the best hyperparameters. Exact details regarding the result can be found in \ref{ssec:effect}.

\subsection{Baselines}
We compare our approach with the following baselines:

\begin{itemize}
\item Transformer: The vanilla Transformer model without any data augmentation.~\cite{vaswani2017attention};
\item Swap: Randomly choose several tokens in the sentence and swap their positions.~\cite{artetxe2017unsupervised};
\item WordDrop: Randomly drop each token in the sentence with probability $\gamma$.~\cite{lample2018unsupervised};
\item Switchout: Randomly replace tokens over the vocabulary by position.~\cite{wang-etal-2018-switchout};
\item SeqMix: Randomly sample two sentences from training dataset and softly combine them.~\cite{guo-etal-2020-sequence}.
\end{itemize}

For Swap and WordDrop, we set the probability $\gamma$ = 0.15 of each word token to be swapped or replaced in the training phase. As for Switchout, both the source and target side temperature parameters are set to 1.0 by us. When adopting the SeqMix training strategy, we set $\beta$ of Beta($\beta$, $\beta$) distribution as 1.0, 1.0 and 0.1 for Zh-En, De-En and En-De respectively. For AdMix, we fix $\beta$ = 1.0 of the Beta($\beta$, $\beta$) distribution and $\alpha$ = 1.0 of the Dirichlet($\alpha$, . . . , $\alpha$) distribution in all experiments.

\subsection{Main Results}
We show the BLEU scores of different methods for Zh-En, En-De, and De-En translation tasks in Table \ref{tab:tab1}. We compare our method with vanilla Transformer and existing methods including Swap, WordDrop, Switchout, SeqMix. AdMix signiﬁcantly outperforms the Transformer baseline on all tasks. Specifically, compared to the baseline system, AdMix delivers significant improvements 1.70, 0.96, and 2.67 points for the Zh-En, En-De, and De-En respectively. The experimental result also shows that our method is more effective on small datasets.

AdMix shows performance advantages over other data augmentation methods, outperforming the best-in-class SwitchOut by 0.8 points in Zh-En and SeqMix by 1.61 points in De-En. In particular, the superiority of AdMix over SeqMix~\cite{guo-etal-2020-sequence} veriﬁes that we propose a more effective method to generate synthetic examples in NMT. Our approach consistently outperforms SeqMix~\cite{guo-etal-2020-sequence}, yielding signiﬁcant 1.03, 0.16, and 1.61 BLEU point gains on the Zh-En, En-De, and De-En translation tasks, respectively. Compared with the method of discrete noise methods (such as WordDrop~\cite{lample2018unsupervised}), our method yields significant 1.39, 0.71, 2.07 BLEU point gains on the Zh-En, En-De, and De-En translation tasks respectively.

\begin{table}[ht]
\renewcommand\tablename{Table}
\centering
\begin{tabular}{lc}\hline
Method  & De-En\\\hline
Transformer & 34.43\\
AdMix  & \textbf{37.10}\\\hdashline
w/o Replace & 37.05\\
w/o Swap    & 36.73\\
w/o Drop    & 37.01\\
w/o Residual Connection  & 36.95\\
Only Source  & 36.97\\
Only Target  & 35.62\\\hline
\end{tabular}
\caption{Ablation study on German-English translation. Only Source and Only Target mean only applying AdMix on the source or target sentences respectively.}\label{tab:tab2}
\end{table}

\subsection{Ablation Study}
To investigate the importance of each component of AdMix, the BLEU scores from the ablation experiments on the German-English dataset are presented in Table \ref{tab:tab2}.
Firstly, we remove one of the three discrete transformation operations which include word replacement, word dropping, and word swapping. And the results appear that no matter which of the three discrete noises is removed, the empirical performance will be damaged. Especially, the BLEU score in the test set drops by 0.37 after removing word swapping. Therefore, it is important to introduce diversity through these three discrete transformation operations. And among these three operations, the word swapping operation can bring more diversity with the same noise levels. We also remove the residual connection to observe the BLEU score, which drops from 37.10 to 36.95 in the test set. The residual connection is effective probably because it preserves the semantic information of clean sentences, allowing a better correspondence between the source and target sentences. 

Finally, We wonder how the BLEU scores would change by only applying AdMix on the source or target sentences. As shown in Table \ref{tab:tab2}, the data augmentation strategies performed on either the source or target sentences are not as effective as on both sides simultaneously. Specially, when we employ AdMix only on the target sentences, the BLEU score drops 1.48 points in the test set.
Nevertheless, both strategies have different degrees of improvement compared to the baseline system. The former brings 2.54 BLEU improvement and the latter yields 1.19 BLEU improvement in the test set.

\begin{figure}[htbp]
\centering
\subfigure[Effect of $\lambda$]{
    \label{fig:fig2a} 
    \includegraphics[width=0.45 \columnwidth]{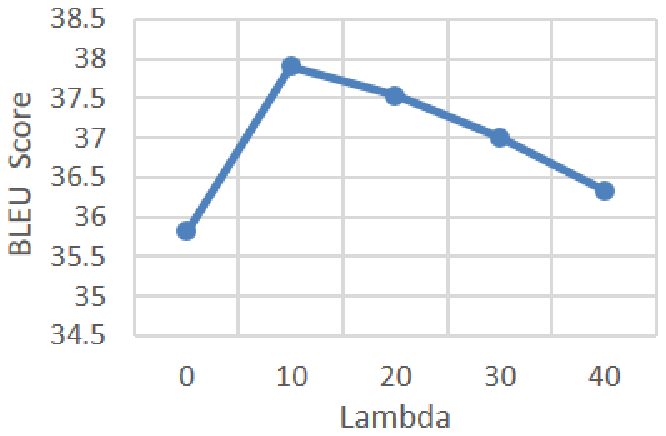}
}
\subfigure[Effect of $\gamma$]{
    \label{fig:fig2b} 
    \includegraphics[width=0.45\columnwidth]{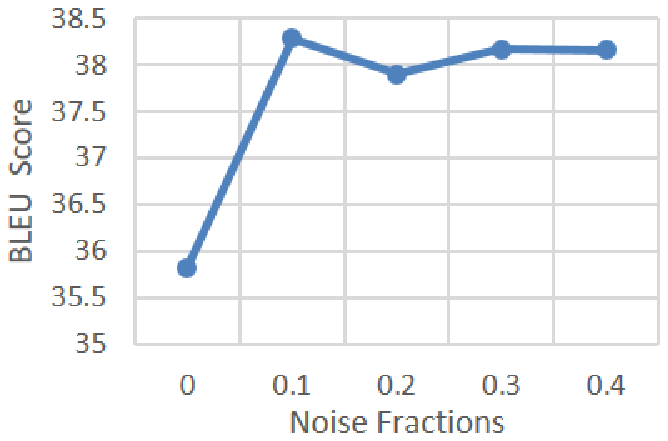}
}
\caption{The effect of lambda and noise fractions on the validation set of German-English translation task.}
\label{fig:fig2}
\end{figure}

\subsection{Effect of $\lambda$ and $\gamma$}
\label{ssec:effect}
We select different values of the hyperparameter $\lambda$ (ranging from 0.0 to 40.0) to investigate the importance of incorporating the Jensen-Shannon (JS) divergence consistency loss. The coefﬁcient of the cross-entropy (CE) loss term is set as 1, and the JS divergence consistency loss term is controlled as the relative weight through the coefficient $\lambda$. The results of the BLEU scores on the German-English dataset are shown in the left panel of Figure \ref{fig:fig2a}. The performance of the model tends to increase and then decrease as the lambda increases, producing the best empirical results on the validation set when the value $\lambda$ is 10. Compared with the standard translation loss, the JS divergence loss can lead to performance improvements. Because it can approximate the original sample output distribution $p(\tilde{y}|x;y)$ and the synthetic sample output distribution $p(\tilde{y}|\hat{x};\hat{y})$ , i.e. to ensure consistency at the semantic level.

Another important hyperparameter with the AdMix approach is the noise fractions $\gamma$, where the fractions can be regarded as the magnitude of perturbations applied to the input sentence. As shown in the right panel of Figure \ref{fig:fig2b}, we apply various noise fractions for AdMix, including 0.1, 0.2, 0.3 and 0.4. It can be observed that the noise fractions of 0.1 shows the best translation performance of the obtained model. Using a larger noise fractions within a certain range can also obtain higher levels of performance than the baseline. The results show that the introduction of discrete noise in the training samples has a positive impact on improving the translation quality of the model, and AdMix can maintain a stable bleu score even at higher noise levels.

\begin{table}[ht]
\renewcommand\tablename{Table}
\centering
\begin{tabular}{l|c|c|c|c}\hline
Method   & Op-0  & Op-1  & Op-2  & Op-3  \\\hline\hline
Transformer & 35.82 & 33.68 & 32.04 & 29.75 \\
Swap     & 35.85 & 33.50 & 31.65 & 29.51 \\
WordDrop & 36.46 & 34.38 & 32.96 & 31.06 \\
Switchout& 36.22 & 34.23 & 32.50 & 30.52 \\
SeqMix   & 36.50 & 34.32 & 32.53 & 30.61 \\\hdashline
AdMix   & \textbf{38.28} & \textbf{36.23} & \textbf{34.52} & \textbf{32.58} \\\hline
\end{tabular}
\caption{Results on German-English validation set. The column
lists results for different noise levels.}\label{tab:tab3}
\end{table}

\subsection{Results to Noisy Inputs}
The noising-based data augmentation methods can not only improve the translation quality of the model, but also improve its robustness~\cite{Miyato2017Adversarial,li2021data}. To test robustness on noisy inputs, similarly to Cheng et al.~\shortcite{cheng-etal-2020-advaug}, we construct a noisy data set by randomly replacing a word in each sentence of the standard German-English validation set with a relevant alternative. The cosine similarity of word embeddings is used to determine the relevance of words.  We repeat the process in an original sentence according to the number of operations where zero operation yields the original clean dataset. 

As shown in Table \ref{tab:tab3}, we can find that our training approach consistently outperforms all baseline methods on all the numbers of operations. This demonstrates that our approach has the ability to resist perturbations. The performance on Transformer baseline decreases rapidly as the number of operations increases. Although the performance of our approach also drops, we can see that our approach consistently exceeds baseline. After conducting three operations on the original validation set, the vanilla Transformer decreases from 35.82 to 29.75, while AdMix decreases from 38.28 to 32.58.

\begin{table}[ht]
\renewcommand\tablename{Table}
\centering
\begin{tabular}{lc}
\hline
Method  &  De-En\\\hline
Transformer  &  34.43\\
+BT & 35.70\\\hdashline
AdMix  & 37.10\\
+BT & \textbf{37.46}\\
\hline
\end{tabular}
\caption{The results of our method combined with back-translation.}\label{tab:tab4}
\end{table}

\subsection{Results with Back-translation}
To explore the effect of combining our method with back-translation (BT), we sample 0.16M English sentences from the WMT13 english monolingual corpus\footnote{https://www.statmt.org/wmt13/training-monolingual-news-2007.tgz}. We compare AdMix against BT on De-En translation task. As shown in Table \ref{tab:tab4}, with using extra monolingual data, the back-translation approach yields a boost of 1.27 BLEU compared to the vanilla Transformer. But the gain delivered by BT is less signiﬁcant than the gain delivered by AdMix. In addition, AdMix and back translation are not mutually exclusive, and we can apply AdMix on the pseudo-parallel sentences obtained by BT to further improve the BLEU score. When we combine BT with AdMix, it will yield 0.36 BLEU improvement than using AdMix alone.

\section{Conclusion}
For machine translation, we propose a new data augmentation method, {\it AdMix}, which performs augmentation in both source and target sentences. We generate new synthetic samples by softly mixing the original sentences with their discrete transformed sentences. Our approach guarantees semantic invariance at both the source and the target sentences compared to the noising-base methods, so it enables the model to focus more on learning the correspondence between the source and target sentences. In the experiment, AdMix delivers improvements over translations tasks at different scales. Experimental results on Chinese-English, German-English, and English-German translation tasks demonstrate the capability of our approach to improve both translation performance and robustness. In the future, besides focusing on the machine translations task in this paper, we are interested in exploring the application of natural language understanding tasks, such as the GLUE benchmark.

\section*{Acknowledgments}
We would like to thank the anonymous reviewers for the helpful comments. This work was supported by Project Funded by the Priority Academic Program Development of Jiangsu Higher Education Institutions.


\bibliographystyle{named}
\end{document}